\documentclass[10pt,twocolumn,letterpaper]{article}

\usepackage{iccv}
\usepackage{times}
\usepackage{epsfig}
\usepackage{graphicx}
\usepackage{amsmath}
\usepackage{amssymb}

\usepackage[ruled]{algorithm2e} %
\usepackage{booktabs} %
\usepackage{graphicx}
\usepackage{makecell}
\usepackage{multirow}
\usepackage{wrapfig}

\usepackage{color}
\usepackage{steinmetz}
\usepackage{graphicx}
\usepackage{amssymb}
\usepackage{amsmath}
\usepackage{mathtools}
\usepackage{algorithmic}
\usepackage{newlfont} %
\usepackage{fixltx2e}
\usepackage[english]{babel} %
\usepackage{enumitem}
\usepackage{microtype}
\usepackage{floatflt}
\usepackage{booktabs} %
\usepackage{subcaption}
\usepackage{epsfig}
\usepackage{xcolor}
\usepackage{tabularx}
\usepackage{ragged2e}

\newcommand{\argmax}{\operatornamewithlimits{arg\,max}}

\newcommand{\mypara}[1]{\vspace{-4mm}\paragraph{#1}}

\usepackage[pagebackref=true,breaklinks=true,letterpaper=true,colorlinks,bookmarks=false]{hyperref}

 \iccvfinalcopy %

\def\iccvPaperID{5314} %

\ificcvfinal\fi
\begin{document}

\title{Block Annotation: Better Image Annotation with Sub-Image Decomposition}

\author{Hubert Lin\\ {\small Cornell University}
\and Paul Upchurch\\{\small Cornell University}
\and Kavita Bala\\{\small Cornell University}
}

\maketitle

\begin{figure*}[t]
  \centering
  \begin{subfigure}[t]{0.25\linewidth}
      \includegraphics[width=\linewidth]{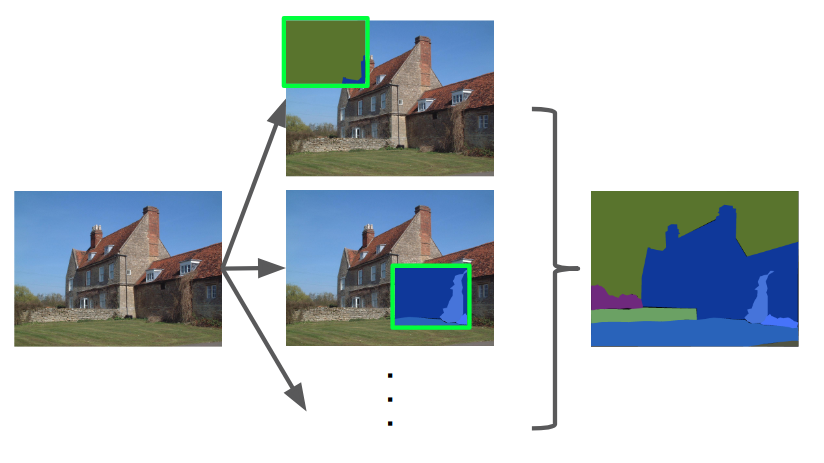}
  \end{subfigure}
  \vline
  \begin{subfigure}[t]{0.30\linewidth}
      \includegraphics[width=\linewidth]{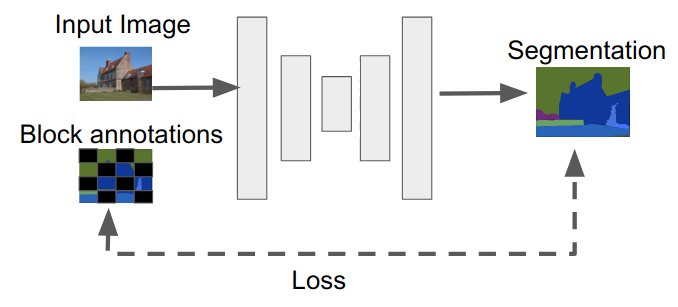}
  \end{subfigure}
  \vline
  \begin{subfigure}[t]{0.30\linewidth}
      \includegraphics[width=\linewidth]{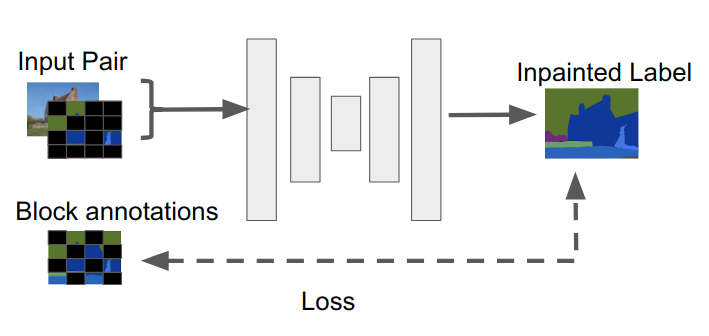}
  \end{subfigure}
  \vspace{-1em}
  \caption{\small (a) Sub-image block annotations are more effective to gather
  than full-image annotations (b) Training on sparse block annotations enables
semantic segmentation performance equivalent to full-image annotations (c) Block
labels can be inpainted with high-quality labels.} \label{fig:teaser}
  \vspace{-2mm}
\end{figure*}

\begin{abstract}

Image datasets with high-quality pixel-level annotations are valuable for
semantic segmentation: labelling every pixel in an image ensures that rare
classes and small objects are annotated. However, full-image annotations are
expensive, with experts spending up to 90 minutes per image. We propose block
sub-image annotation as a replacement for full-image annotation. Despite the
attention cost of frequent task switching, we find that block annotations can be
crowdsourced at higher quality compared to full-image annotation with equal
monetary cost using existing annotation tools developed for full-image
annotation.  Surprisingly, we find that 50\% pixels annotated with blocks allows
semantic segmentation to achieve equivalent performance to 100\% pixels
annotated.  Furthermore, as little as 12\% of pixels annotated allows
performance as high as 98\% of the performance with dense annotation. In
weakly-supervised settings, block annotation outperforms existing methods by
3-4\% (absolute) given equivalent annotation time. To recover the necessary
global structure for applications such as characterizing spatial context and
affordance relationships, we  propose an effective method to inpaint
block-annotated images with high-quality labels without additional human effort.
As such, fewer annotations can also be used for these applications compared
to full-image annotation.

\end{abstract}

\section{Introduction}
\label{section:introduction}

Recent large-scale computer vision datasets place a heavy emphasis on
high-quality fully dense annotations (in which over 90\% of the pixels are
labelled) for hundreds of thousands of images. Dense annotations are
valuable for both semantic segmentation and applications beyond segmentation
such as characterizing spatial context and affordance relationships
\cite{cocostuff, hassanin2018visual}. The
long-tail distribution of classes means it is difficult to gather annotations
for rare classes, especially if these classes are difficult to segment.
Annotating every pixel in an image ensures that pixels corresponding to rare
classes or small objects are labelled. Dense annotations also capture pixels
that form the boundary between classes. For applications such as understanding
spatial context between classes or affordance relationships, dense annotations
are required for principled conclusions to be drawn.  In the past, polygon
annotation tools have enabled partially dense annotations (in which small
semantic regions are densely annotated) to be crowdsourced at scale with public
crowd workers. These tools paved the way for the cost-effective creation of
large-scale partially dense datasets such as ~\cite{bell13opensurfaces, mscoco}.
Despite the success of these annotation tools, fully dense datasets have relied
extensively on expensive expert annotators~\cite{bdd100k, cityscapes,
pascalcontext, ade20k, mapillary} and private crowdworkers~\cite{cocostuff}.

We propose annotation of small blocks of pixels as a stand-in replacement for
full-image annotation (figure~\ref{fig:teaser}). We find that these annotations
can be effectively gathered by crowdworkers, and that annotation of a sparse
number of blocks per image can train a high performance segmentation network.
We further show these sparsely annotated images can be extended automatically to
full-image annotations.

We show block annotation has:
\begin{itemize}[wide=0pt]
    \vspace{-2mm}
\item \textbf{Wide applicability.} (Section \ref{sec:block_anno})
Block annotations can be effectively crowdsourced at higher quality compared to
full annotation.  It is easy to implement and works with existing advances in
  image annotation.   \vspace{-2mm}
\item  \textbf{Cost-efficient Design.} (Section \ref{sec:block_anno})
  Block annotation reflects a cost-efficient design paradigm (while current
  research focuses on reducing annotation time).  This is reminiscent of
  gamification and citizen science where enjoyable tasks lead to low-cost
  high-engagement work.  \vspace{-2mm}
\item \textbf{Complex Region Annotation.} (Section \ref{sec:block_anno})
    Block annotation shifts focus from categorical regions to spatial
    regions. When annotating categorical regions,
    workers segment simple objects before complex objects. With
    spatial regions, informative complex regions are forced to be annotated.
    \vspace{-2mm}
  \item \textbf{Weakly-Supervised Performance.} (Section \ref{sec:segmentation}) Block annotation is competitive in weakly-supervised settings, outperforming existing methods by 3-4\% (absolute) given equivalent
  annotation time.
  \vspace{-2mm}
\item \textbf{Scalable Performance.} (Section \ref{sec:segmentation})
  Full-supervision performance is achieved by annotating 50\% of blocks per
  image. Thus, blocks can be annotated until desired performance is achieved,
  in contrast to methods such as scribbles.  \vspace{-2mm}
\item  \textbf{ Scalable Structure.} (Section \ref{sec:block_to_dense})
    Block-annotated images can be effectively inpainted with high quality labels
    without additional human effort.
\vspace{-2mm}
\end{itemize}

\section{Related Work}
\label{sec:related}

In this section we review recent works on pixel-level annotation in three areas:
human annotation, and human-machine
annotation, and dense segmentation with weak supervision.

\mypara{Human Annotation.} Manual labeling of every pixel is impractical for
large-scale datasets. A successful method is to have crowdsource workers segment
polygonal regions to click on boundaries.  Employing crowdsource workers offers
its own set of challenges with quality control and task
design~\cite{cubam,bell13opensurfaces,cag}. Although large-scale public
crowdsourcing can be successful~\cite{mscoco} recent benchmark datasets have
resorted to in-house expert annotators ~\cite{cityscapes,mapillaryvistas}.
Annotation time can be reduced through improvements such as autopan,
zoom~\cite{bell13opensurfaces} and shared polygon boundaries~\cite{bdd100k}.
Polygon segmentation can be augmented by painted labels on superpixel
groups~\cite{cocostuff} and Bezier curves~\cite{bdd100k}. Pixel-level labels for
images can also be obtained by (1) constructing a 3D scene from an image
collection, (2) grouping and labeling 3D shapes and (3) propagating shape labels
to image pixels~\cite{labelfusion}.  In our work, we investigate sub-image
polygon annotation, which can be further combined with other methods (sec.
\ref{sec:block_anno}.)

\mypara{Human-Machine Annotation.}

Complex boundaries are time-consuming to trace manually.  In these cases the
cost of pixel-level annotation can be reduced by automating a portion of the
task.  Matting and object selection~\cite{grabcut, levin2008closed,
  levin2008spectral, bai2009geodesic, xu2016deep, xu2017deep,
boroujerdi2017deep, le2018interactive, xu2017matt} generate tight boundaries
from loosely annotated boundaries or few inside/outside clicks and scribbles.
\cite{extremeclicks,dextr} introduced a predictive method which automatically
infers a foreground mask from 4 boundary clicks, and was extended to full-image
segmentation in \cite{agustsson2019interactive}. The number of boundary clicks
was further reduced to as few as one by~\cite{polygonrnn}.  Predictive methods
require an additional human verification step since the machine mutates the
original human annotation. The additional step can be avoided with an online
method.  However, online methods (e.g., \cite{polygonrnn, le2018interactive,
agustsson2019interactive}) have higher requirements since the algorithm must be
translated into the web browser setting and the worker's machine must be
powerful enough to run the algorithm\footnote{Offloading online methods onto a
cloud service offers a different landscape of higher costs (upfront development
and ongoing operation costs).}.  Alternatively, automatic proposals can be
generated for humans to manipulate: \cite{fluid} generates segments, \cite{sss}
generates a set of matting layers, \cite{zhang2018collaborative} generates
superpixel labels, and \cite{qin2018bylabel} generates boundary fragments. In
our work, we show that human-annotated blocks can be extended automatically into
dense annotations (sec. \ref{sec:block_to_dense}), and we discuss how other
human-machine methods can be used with blocks (sec. \ref{sec:compat}).

\mypara{Weakly-Supervised Dense Segmentation.}

There are alternatives to training with high-quality densely annotated images
which substitute quantity for label quality and/or richness. Previous works have
used low-quality pixel-level annotations~\cite{zlateski2018importance}, bounding
boxes~\cite{pap2015weakly, khoreva2017simple, copypaste},
point-clicks~\cite{bearman2016what}, scribbles~\cite{bearman2016what,
lin2016scribblesup}, image-level class labels~\cite{pap2015weakly,
shi2017weakly, ahn2018learning}, image-level text
descriptions~\cite{hong2017weakly} and unlabeled related web
videos~\cite{hong2017weakly} to train semantic segmentation networks. Combining
weak annotations with small amounts of high-quality dense annotation is another
strategy for reducing cost~\cite{minc, hu2017learning}.
\cite{shen2018bootstrapping} proposes a two-stage approach where image-level
class labels are automatically converted into pixel-level masks which are used
to train a semantic segmentation network. We find a small number of sub-image
block annotations is a competitive form of weak supervision (sec.
\ref{sec:weak_comp}).

\section{Block Annotation \label{sec:block_anno}}

Sub-image block annotation is composed of three stages: (1) Given an image $I$,
select a small spatial region $I'$; (2) Annotate $I'$ with
pixel-level labels; (3) Repeat (with different $I'$) until $I$ is sufficiently
annotated. In this paper, we explore the case where $I'$ is
rectangular, and focus on the use of existing pixel-level annotation tools.

Can block annotations be gathered as effectively as full-image annotations with
existing tools? In section \ref{sec:annotation_interface}, we show our
annotation interface. In section \ref{sec:block_quality}, we explore the quality
of block annotation versus full-image annotation.  In section
\ref{sec:block_viability}, we examine block annotation for a real-world dataset.
In section \ref{sec:worker_feedback}, we discuss the cost of block annotation
and show worker feedback.  In section \ref{sec:block_selection}, we discuss how
blocks for annotation can be selected in practice. Finally, in section
\ref{sec:compat} we discuss the compatibility of block annotation with existing
annotation methods.

\subsection{Annotation Interface}
\label{sec:annotation_interface} 
\begin{figure}[t!]
  \centering
  \begin{subfigure}[t]{0.49\linewidth}
      \includegraphics[width=\linewidth]{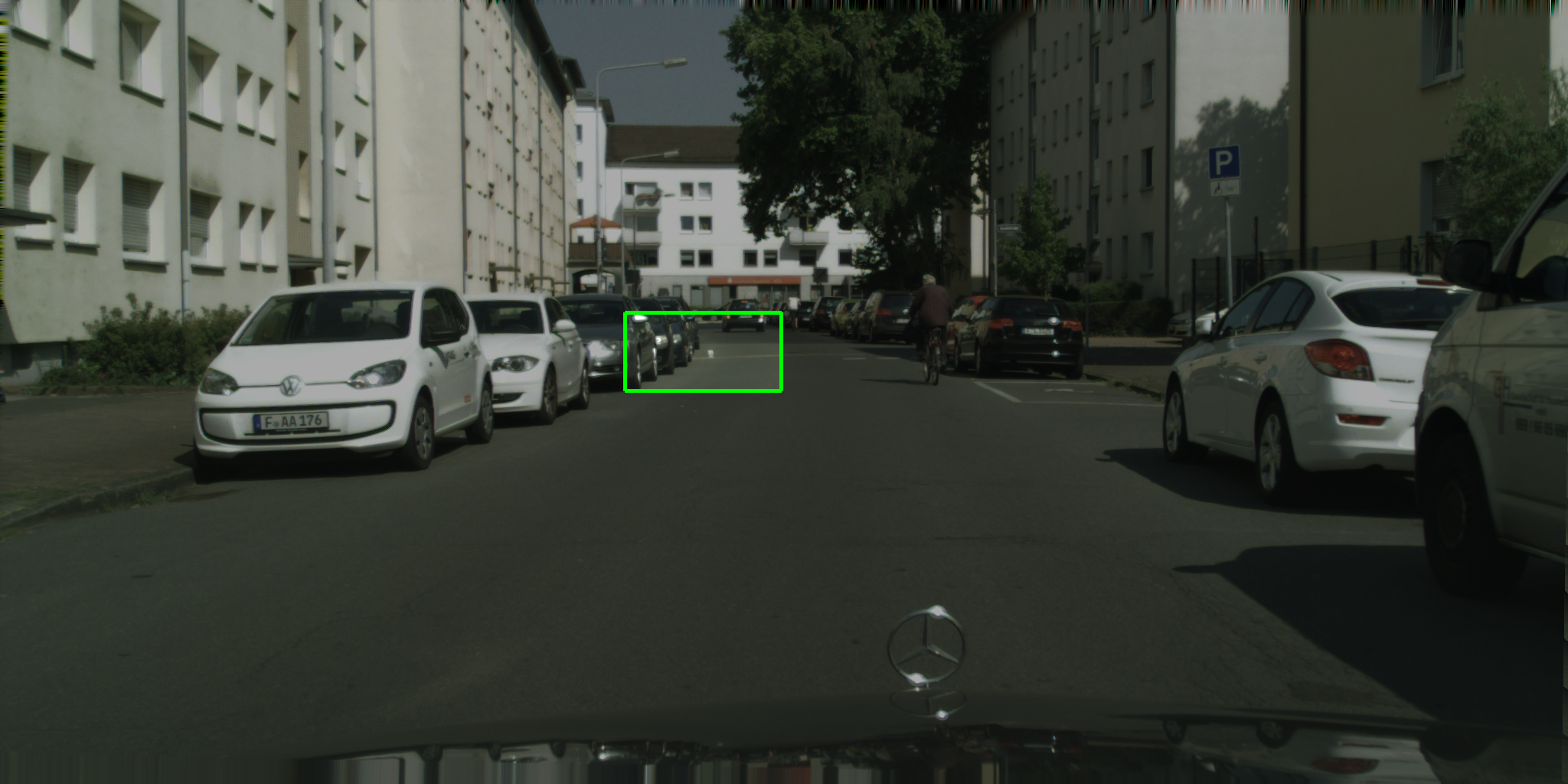}
      \caption{\small Highlighted block.}
  \end{subfigure}
  \begin{subfigure}[t]{0.49\linewidth}
      \includegraphics[width=\linewidth]{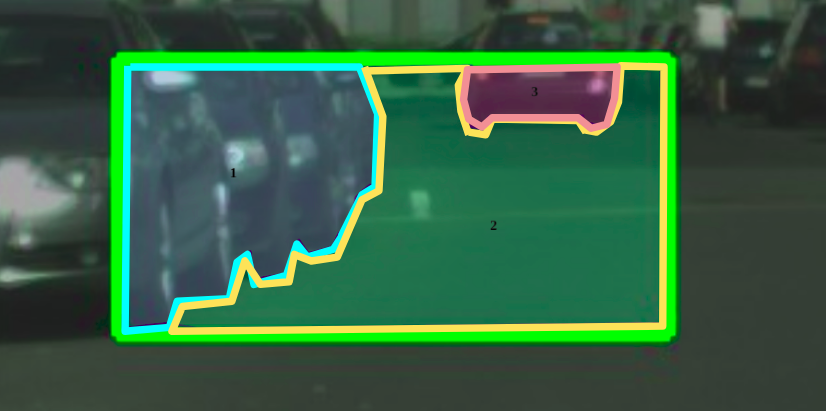}
      \caption{\small Finished block annotation.}
  \end{subfigure}
  \vspace{-2mm}
  \caption{\small \textbf{Block Annotation UI}. Annotators are given one
highlighted block to annotate with the remainder of the image as context.}
  \label{fig:annotation_interface}
\end{figure}

Our block annotation interface is given in
figure \ref{fig:annotation_interface} and implemented with existing tools
~\cite{bell13opensurfaces}. For full image annotation, the highlighted block
covers the entire image. Studies are deployed on Amazon Mechanical Turk.

\subsection{Quality of Block Annotation}
\label{sec:block_quality}

\begin{figure}[t!]
  \centering
  \begin{subfigure}[t]{0.49\linewidth}
    \includegraphics[width=\linewidth]{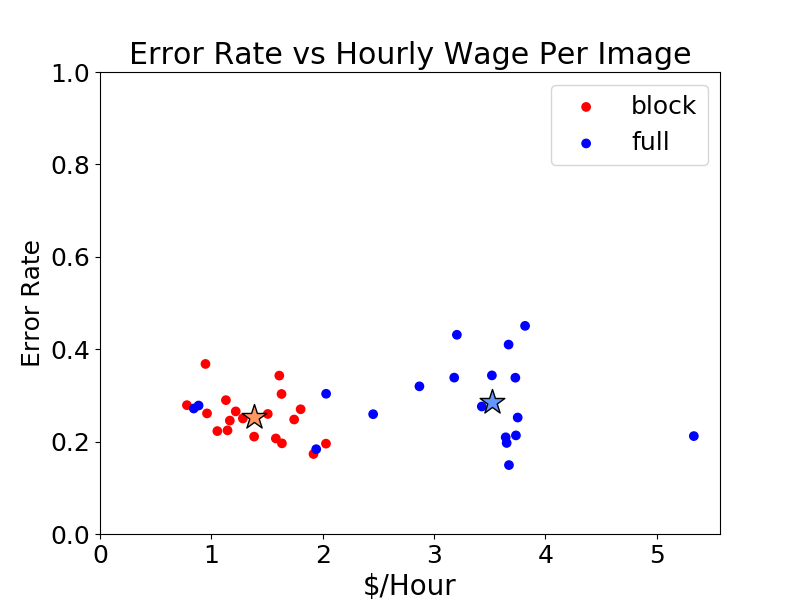}
    \caption{\small}
  \end{subfigure}
  \begin{subfigure}[t]{0.49\linewidth}
    \includegraphics[width=\linewidth]{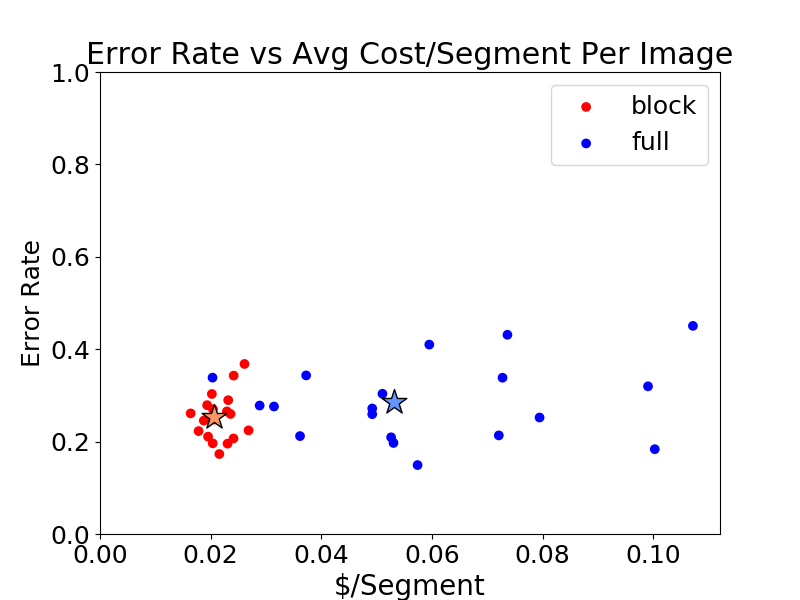}
    \caption{\small}
  \end{subfigure}
\vspace{-1em}
  \caption{\small \textbf{Annotation error rate} for block and full
    annotation. Each point represents one image. The same set of images are both
    block annotated and full-image annotated. The stars represent the centroid
    (median). Cost/time include estimated cost/time to assign labels for each
    segment \cite{bell13opensurfaces}. \textbf{Lower-left is better}. With
    block annotation, workers (a) choose to work for lower wages and (b) segment
    more regions for less pay per region. The overall quality is higher for
    block annotation.}.
    \label{fig:quality_results}
    \vspace{-2mm} 
\end{figure}

\begin{figure}[t!]
  \centering

  \begin{subfigure}[t]{0.31\linewidth}
  \includegraphics[width=\linewidth]{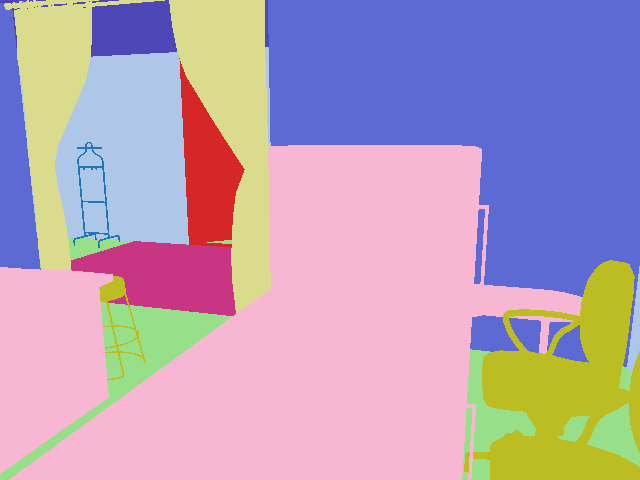}
  \end{subfigure}
  {\unskip\ \vrule\ }
  \begin{subfigure}[t]{0.31\linewidth}
  \includegraphics[width=\linewidth]{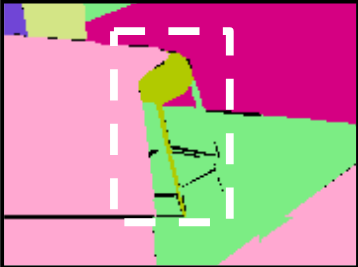}
  \end{subfigure}
  \begin{subfigure}[t]{0.31\linewidth}
  \includegraphics[width=\linewidth]{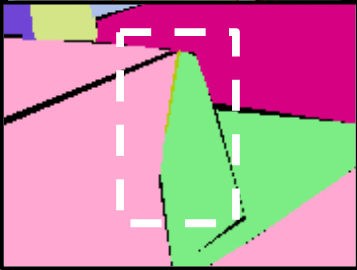}
  \end{subfigure}

  \vspace{-1em} \caption{\small \textbf{SUNCG/CGIntrinsics annotation}.  (a)
Ground truth. (b) Block annotation (zoomed-in) (c) Full annotation (zoomed-in).
White dotted box highlights an example where block annotation qualitatively
outperforms full annotation.  More in supplemental.}
\vspace{-2mm}
\label{fig:block_small_ex}
\end{figure}

We explore the quality of block annotations compared to full-image annotations
on a synthetic dataset. How does the quality and cost compare between block and
full annotations? We find that \emph{the average quality for block-annotated
  images is higher while the total monetary cost is about the same}. The average
  quality of block annotations is consistently higher including for small
  regions (e.g. fig \ref{fig:block_small_ex}). The overall block annotation
  error is 12\% lower than full annotation. For regions smaller than 0.5\% of
  the image, the block annotation error is 6\% lower.  In figure
  \ref{fig:quality_results}, the cost and quality of block versus full image
  annotation is shown. Remarkably, we find that \emph{workers are willing to
    work on block annotation tasks for a significantly lower hourly wage.  This
  indicates that block annotation is more intrinsically palatable for
crowdworkers}, in line with \cite{huang2010toward} which shows task design can
influence quality of work.  
Moreover, workers are more likely to over-segment objects with respect
to ground truth (e.g.  individual cushions on a couch, handles on cabinets) with
block annotation tasks. Note that block boundaries may also divide semantic
regions.  Table \ref{table:block_anno_numbers} contains additional statistics.
Despite similar costs to annotate an image in blocks or in full, we show in
section \ref{sec:segmentation} that competitive performance is achieved with
less than half of the blocks annotated per image.

\begin{table}[t!]
 \small
 \begin{tabularx}{\linewidth}{|c|X|X|}
   \hline
   \makecell[l]{} &
   \makecell[l]{Block } &
   \makecell[l]{Full} \\
   \hline
   \hline
   \makecell[l]{Error } & 0.253 & 0.286  \\
   \makecell[l]{Error (small regions)} & 0.636 & 0.677 \\
   \hline
   \makecell[l]{\$ / hr } & \$1.40 / hr & \$3.12 / hr  \\
   \hline
   \makecell[l]{Total cost} & \$2.00 & \$2.05  \\
   \makecell[l]{Total cost (median)} & \$1.99 & \$2.23  \\
   \hline
   \makecell[l]{\# segments} & 95.68 & 38.95  \\
   \makecell[l]{\$ / segment } & \$0.0215 & \$0.0595  \\
   \hline
\end{tabularx}
\vspace{-1em}
\caption{\small \textbf{Block vs Full Annotation}. Average statistics per image.}
\label{table:block_anno_numbers}
\vspace{-2mm}
\end{table}

\mypara{Study Details.}

For these experiments, we chose to use a synthetic dataset.  While human
annotations may contain mistakes, synthetic datasets are generated with known
ground truth labels with which annotation error can be computed. The
CGIntrinsics dataset \cite{li2018cgintrinsics} contains physically-based
renderings of indoor scenes from the SUNCG dataset \cite{suncg, pbrs}. We use
the more realistic CGIntrinsics renderings and the known semantic labels from
SUNCG. The labels are categorized according to the
NYU40\cite{gupta2013perceptual} semantic categories. Due to the nature of indoor
scenes, the depth and field of view of each image is smaller than outdoor
datasets.  The reduced complexity means that crowdworkers are able to produce
good full-image annotations for this dataset.  

We select MTurk workers who are skilled at both full-image annotation and block
annotation in a pilot study (a standard quality control practice
\cite{bell13opensurfaces}). The final pool consists of 10 workers.  Image
difficulty is estimated by counting the maximum number of ground truth segments
in a fixed-size sliding window.  Windows, mirrors, and void regions are masked
out in the images so that workers do not expend effort on visible content for
which ground truth labels do not exist (such as objects seen through a window or
mirror). We manually cull images that include transparent glass tables which are
not visible in the renderings, or doorways through which visible content can be
seen but no ground truth labels exist.  After filtering, twenty of the one
hundred most difficult images are selected. We choose a block size so that an
average of 3.5 segments are in each block. This results in 16 blocks per image.
For each task, a highlighted rectangle outlines the block to be annotated. We
find that workers will annotate up to the inner edge of the highlighted
boundary. Therefore, we ensure the edges of the rectangle do not overlap with
the region to be annotated.

Workers are paid \$0.06\footnote{\$ refers to USD in throughout this paper.} per
block annotation task and \$0.96 per full-image task. Bonuses up to 1.5 times
the base pay are awarded to attempt to raise the effective hourly wage for
difficult tasks to \$4 / hr.  Our results show that workers are willing to work
on block annotation tasks beyond the time threshold for bonuses, effectively
producing work for an hourly wage significantly lower than the intended \$4 /
hr.  On the other hand, workers do not often exhibit this behavior with full
annotation tasks. Different workers may work on different blocks belonging to
the same image. We use two forms of quality control: (1) annotations must
contain a number of segments greater than 25\% of the known number of
ground-truth segments for that task and (2) annotations cannot be submitted
until at least 10 seconds / 3 minutes (block / full) have passed. All
submissions satsifying these conditions are accepted during the user study.  For
an overview of QA methods, please refer to \cite{quinn2011human}.  Labels are
assigned by majority ground-truth voting, with cost estimated from
\cite{bell13opensurfaces}.

To evaluate the quality of annotations in an image with K classes, we measure the class-balanced
error rate (class-balanced Jaccard distance):
\begin{align}
  error\, rate &= \dfrac{1}{K} \sum_{c=1}^{K} \dfrac{(FP_c + FN_c)}{(TP_c + FP_c + FN_c)}\\
\nonumber            &= 1 - mIOU
\end{align}

\subsection{Viability of Real-World Block Annotation}
\label{sec:block_viability}

\begin{figure}[t!]
  \centering
  \includegraphics[width=\linewidth]{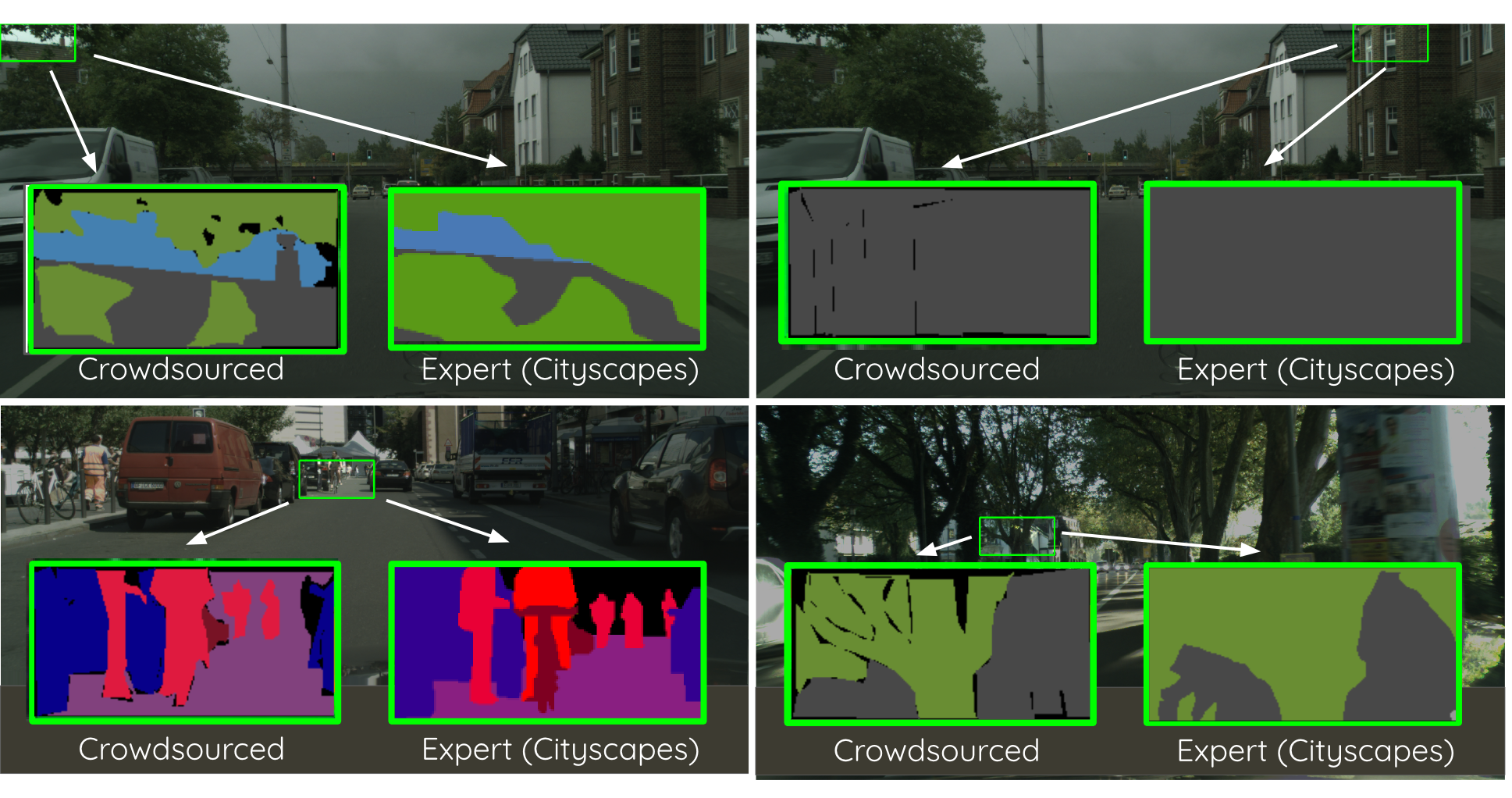}
\vspace{-1em} 

\caption{\small \textbf{Crowdsourced vs expert segments}. Crowdsourced
block-annotated segments are compared to expert Cityscapes segments.
Crowdsourced segments are colored for easier comparison. Top-left is a
high-quality example. See supplemental for more.}
\label{fig:crowd_vs_expert_filled}
\end{figure}

How does block annotation fare with a real-world non-synthetic dataset?  To
study the viability of block annotating real-world datasets with scalable
crowdsourcing, we ask crowdworkers to annotate blocks from images in
Cityscapes~\cite{cityscapes}. We choose Cityscapes for the annotation complexity
of its scenes -- 1.5 hours of expert annotation effort is required per image. In
contrast, other datasets such as \cite{pascalcontext, cocostuff} require less
than 20 minutes of annotation effort per image. We expect crowd work to be worse
than expert work, so it is a surprisingly positive result that the quality of
the crowdsourced segments are visually comparable to the expert Cityscapes
segments (figure \ref{fig:crowd_vs_expert_filled}).  Some crowdsourced segments
are very high quality. %
We find that 47\% of blocks
have more crowd segments than expert segments (20\% have fewer segments, and
33\% have the same \# of segments). A summary of the cost is given in table
\ref{table:crowd_vs_expert_cost} which compares public crowdworkers to trained
experts. It is feasible block annotation time will decrease with expert
training. Given 100 uniformly sized blocks per image, we ask an expert to create
equal-quality block and full annotations; we find one block is 1.56\% of the
effort of a full image.

\begin{table}[t!]
 \small
 \begin{tabularx}{\linewidth}{|c|X|X|}
   \hline
   \makecell[l]{} & 
   \makecell[l]{Block (Crowd)} &
   \makecell[l]{Full (Expert \cite{cityscapes, mapillary}}) \\
   \hline
   \makecell[l]{\$ / Task} & \$0.13 &  - \\
   \makecell[l]{Time / Task} & 2 min & 1.5 hr  \\
   \hline
\end{tabularx}
\vspace{-1em}
\caption{\small \textbf{Real-world cost of annotation.} Cost evaluated on
Cityscapes. Each block is annotated by MTurk workers. Full-image is annotated by
experts in \cite{cityscapes}. Note: \cite{cityscapes} annotates instance
segments. See table \ref{table:block_anno_numbers} for crowd-to-crowd
comparison.  } 
\vspace{-2mm}
\label{table:crowd_vs_expert_cost} \end{table}

\mypara{Study Details.}
We searched for workers who produce high-quality work in a pilot study and found a
set of 7 workers. These workers were found within a hundred pilot HITs (for a
total cost of \$4). We approved all of their submissions during the user study.
We do not restrict workers from annotating outside of the block, and we do not
force workers to densely annotate the block.  We do not include the use of
sentinels or tutorials as in ~\cite{bell13opensurfaces}.  

Thirteen randomly selected validation images from Cityscapes are annotated by
crowdworkers. Each image is divided into 100 uniformly shaped blocks. A total of
650 (50 per image) are annotated in random order. Workers are paid \$0.06 per
task. Workers are automatically awarded bonuses so that the effective hourly
wage at least \$5 for each block, with bonuses capped at \$0.24 to prevent
abuse.  For one block, the total base payout is \$0.06 with an average of
\$0.0636 in bonuses over 93 seconds of active work.  On average, each annotated
block contains 3.5 segments.  Assigning class labels will cost an additional
\$0.01 and 26 seconds ~\cite{bell13opensurfaces}.  To be consistent with
Cityscapes, we instruct workers to not segment windows, powerlines, or small
regions of sky between leaves. However, workers will occasionally choose to do
so and submit higher quality segments than required.  

\subsection{Annotation Cost and Worker Feedback}
\label{sec:worker_feedback}

Our costs (tables \ref{table:block_anno_numbers},
\ref{table:crowd_vs_expert_cost}) are aligned with existing large-scale studies.
Large-scale datasets \cite{bell13opensurfaces, mscoco} show that cultivating
good workers produces high quality data at low cost.  Table 2 of
\cite{hara2018data} reports a median wage of \$1.77/hr  to \$2.11/hr; the median
MTurk wage in India is \$1.43/hr \cite{hara2019worker}.  For ``image
transcription'', the median wage is \$1.13/hr over 150K tasks.

\emph{Workers gave overwhelmingly positive feedback for block annotation} (table
\ref{table:worker_feedback}), and we found that some workers would reserve
hundreds of block annotation tasks at once.  Only 3 out of the 57 workers who
successfully completed at least one pilot or user study task requested higher
pay.  In contrast, our pilot studies showed that \emph{workers are unwilling to
accept full-image annotation tasks if the payment is reduced to match the wage
of block annotation.} We conjecture that task enjoyment leads
to long term high-quality output (c.f. \cite{huang2010toward}).

\begin{table}[th!]
 \small
 \begin{tabularx}{\linewidth}{|c|l|l|l|l|l|X|}
   \hline
   \makecell[l]{} &
   \makecell[l]{\!\!``Nice''\!\!\\\!\!``Good''\!\!\\\!\!``Great''\!\!} &
   \makecell[l]{\!\!``Fun''\!\!\\\!\!``Happy''\!\!} &
   \makecell[l]{\!\!``Easy''} &
   \makecell[l]{\!\!``Okay''\!\!} &
   \makecell[l]{\!\!Release\!\!\\\!\!More HITs\!\!} &
   \makecell[l]{\!\!Increase\!\!\\\!\!Pay\!\!} \\
   \hline
     \makecell[l]{\!\!\#\!\!} & 8 & 5 & 4 &  2 & 2 & 3 \\
   \hline
\end{tabularx}
\vspace{-1em}
\caption{\small \textbf{Block annotation worker feedback}. Free-form responses are
aggregated over SUNCG and Cityscapes experiments, and collected at most once per
worker. 
All 24 sentiments across all 19 worker responses are
summarized. 
}
\label{table:worker_feedback}
\vspace{-3mm}
\end{table}

\subsection{Block Selection}
\label{sec:block_selection}

Our experiments show that workers are comfortable annotating between 3 to 6
segments per block. Therefore, block size can be selected by picking a size such
that the average number of segments per block falls in this range. For a novel
dataset, this can be done fully labelling several samples and producing an
estimate from the fully labelled samples. Without priors on spatial distribution
of rare classes or difficult samples within an image, a checkerboard or
pseudo-checkerboard pattern of blocks focuses attention (across different tasks)
uniformly across the image. Far apart pixels within an image are less
correlated than neighboring pixels. Therefore, it is good to sample
blocks that are spread out to encourage pixel diversity within images.

\subsection{Compatibility with Existing\\ Annotation Methods}
\label{sec:compat} 

Block annotation is compatible with many annotation tools and
innovations besides polygon boundary annotation.

\mypara{Point-clicks and Scribbles.}

Annotations such as point clicks or scribbles are faster to acquire than
polygons, which leads to a larger and more varied dataset at the same cost.
Combining this with blocks will further increase annotation variety due to the
diversity that come from annotating a few blocks in many images over annotating
fewer number of images fully.  Additionally, ~\cite{bearman2016what, minc} show
that the most cost-effective method for semantic segmentation is a combination
of densely annotated images and a large number of point clicks.  The densely
annotated images can be replaced by polygon block annotations since they also
contain class boundary supervision for the segmentation network.

\mypara{Superpixels.} Superpixel annotations enable workers to mark a group of
visually-related pixels at once~\cite{cocostuff}. This can reduce the annotation
time for background regions and objects with complex boundaries. Superpixel
annotation can be easily deployed to our block annotation setting.

\mypara{Polygon Boundary Sharing.} Boundary sharing reuses existing
boundaries so that workers do not need to trace each boundary
twice~\cite{bdd100k}. This approach can be easily deployed in our block
annotation setting.

\mypara{Curves.} Bezier tools allow workers to quickly annotate
curves~\cite{bdd100k}.  It can be easily deployed in our block annotation
setting but it may be less effective on long curves since each part of
the curve must be fit separately.

\mypara{Interactive Segmentation.} Recent advances in interactive segmentation
(e.g., ~\cite{polygonrnn, dextr, agustsson2019interactive}) utilize neural
networks to convert sparse human inputs into high quality segments. For novel
domains without large-scale training data, block-annotated images can act as
cost-efficient seed data to train these models. Once trained, these methods can
be applied directly to each block, although further analysis should be conducted
to explore the efficiency of such an approach due to block boundaries splitting
semantic regions.

\section{Segmentation Performance \label{sec:segmentation}}

How well do block annotations serve as training data for semantic segmentation?
In section~\ref{sec:setup}, the experimental setup is summarized.  In
section~\ref{section:evaluation}, we evaluate the effectiveness of block
annotations for semantic segmentation. In section ~\ref{sec:weak_comp}, we
compare block annotation with existing weakly supervised segmentation methods.

\subsection{Experimental Setup}
\label{sec:setup}

\paragraph{Pixel Budget.} We vary the ``pixel budget'' in our experiments to
explore segmentation performance across a range available annotated pixels.
``Pixel budget'' refers to the \% of pixels annotated across the training
dataset, which can be controlled by varying the number of annotated images, the
number annotated blocks per image, and the size of blocks per image. Our block
sizes are fixed in our experiments.

\mypara{Block Size.} We divide images into a 10-by-10 grid for our
experiments.

\mypara{Block Selection.}

We experiment with two block selection strategies: (a) Checkerboard annotation
and (b) Pseudo-checkerboard annotation. Checkerboard annotation means that every other block in
a variable number of images are annotated. Pseudo-checkerboard annotation means that every N blocks
are annotated in every image, where N is $\dfrac{\texttt{\# pixels in
dataset}}{\texttt{pixel budget}}$.  For example, with a pixel budget equivalent
to 25\% of the dataset, every fourth block is annotated for the entire dataset.
At pixel budget 50\%, checkerboard and pseudo-checkerboard are identical.

For the remainder of the paper, ``Block-X\%'' refers to pseudo-checkerboard
annotation in which X\% of the blocks per image are annotated.

\mypara{Sementation Model.} We use DeepLabv3+\cite{deeplabv3plus2018} initialized with the official
pretrained checkpoint (pretrained on ImageNet~\cite{deng2009imagenet} +
MSCOCO~\cite{mscoco} + Pascal VOC~\cite{everingham10pascal}). The network is
trained for a fixed number of epochs. See supplemental for additional details.

\mypara{Datasets.} Cityscapes is a dataset with ground truth annotations for 19
classes with 2975 training images and 500 validation images. ADE20K contains
ground truth annotations for 150 classes with 20210 training images and 2000
validation images. These datasets are chosen for their high quality dense ground
truth annotations and for their differences in  number of images / classes and
types of scenes represented. The block annotations are synthetically generated
from the existing annotations.

\subsection{Evaluation}
\label{section:evaluation}

\paragraph{Blocks vs Full Image.}

\begin{figure}[t!]
  \centering
  \includegraphics[width=0.95\linewidth]{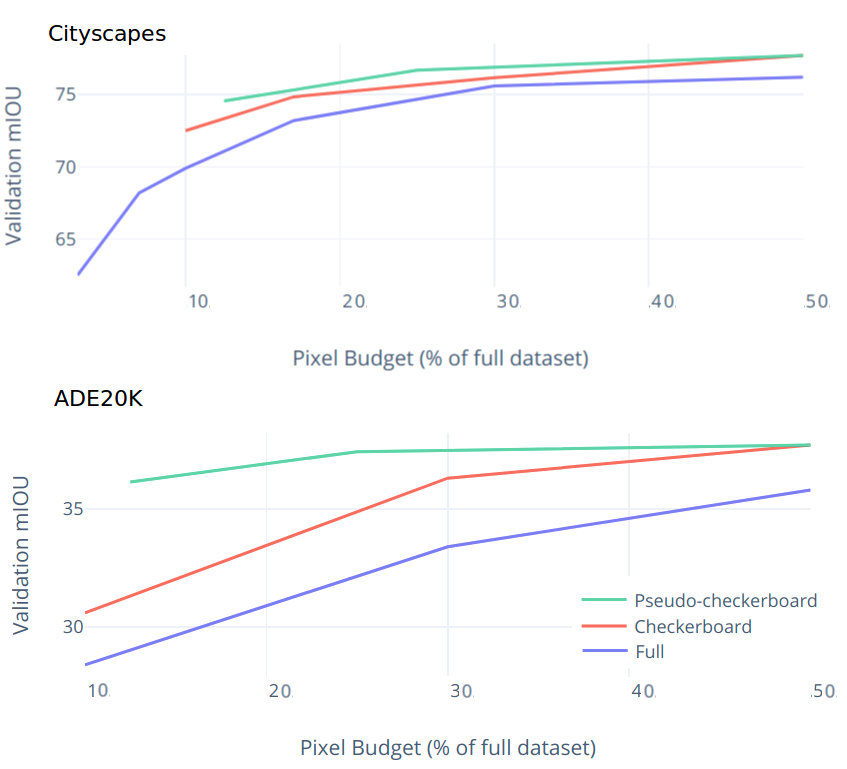}
\vspace{-1em} \caption{\small \textbf{Semantic segmentation performance}.
Training images are annotated with different pixel budgets. Pseudo-checkerboard
block annotation outperforms checkerboard and full annotation.} \label{fig:budget_acc}
\vspace{-2mm}
\end{figure}

How does block annotation compare to full-image annotation for semantic
segmentation? We plot the mIOU achieved when trained on a set of annotations
against pixel budget in figure \ref{fig:budget_acc}. 

For both Cityscapes and ADE20K, \emph{block
annotation significantly outperforms full-image annotation}.  The performance
gap widens as the pixel budget is decreased -- at pixel budget 12\%, the
reduction in error from full annotation to block annotation is 13\% (10\%)
Cityscapes (ADE20K).  
Our results indicate that the quantity of annotated images is more valuable than
the quantity of annotations per image. The pseudo-checkerboard block selection
pattern consistently outperforms the checkerboard block selection pattern and
full annotation. For any pixel budget, pseudo-checkerboard block annotation
annotates fewer pixels per image which means more images are annotated.

\begin{table}[ht!]
 \small
 \begin{tabularx}{\linewidth}{|c|X||X|X|}
   \hline
   \makecell[l]{} & 
   \makecell[l]{\!\!Optimal (Full)\!\!} &
   \makecell[l]{\!\!Block-50\%\!\!} &
   \makecell[l]{\!\!Block-12\%\!\!} \\
   \hline
   \makecell[l]{\!\!Cityscapes} & 77.7 & 77.7 & 74.6 \\
   \makecell[l]{\!\!ADE20K} & 37.4 & 37.2 & 36.1 \\
   \hline
\end{tabularx}
\vspace{-1em} 
\caption{\small \textbf{Semantic segmentation performance} when trained on all
images.  Training with block annotations uses fewer annotated pixels than full
annotation but achieves equivalent performance.  } \label{table:block_vs_opt}
\vspace{-2mm}
\end{table}

\mypara{Blocks vs Optimal Performance.}

How many blocks need to be annotated for segmentation performance to approach
the performance achieved by training on full-image annotations for the entire
dataset? In table \ref{table:block_vs_opt}, we show results when the network is
trained on the full dataset compared to pseudo-checkerboard blocks. Remarkably, we
find that \emph{checkerboard blocks with 50\% pixel budget allow the network to
achieve similar performance to the full dataset with 100\% pixel budget},
indicating that at least 50\% of the pixels in Cityscapes and ADE20K are
redundant for learning semantic segmentation.  Furthermore, with only 12\% of
the pixels in the dataset annotated, relative error in segmentation performance
is within 12\%/2\% of the optimal for Cityscapes/ADE20K. These results suggest
that fewer than 50\% of the blocks in an image need to be annotated for training
semantic segmentation, reducing the cost of annotation reported in section
\ref{sec:block_anno}.

\subsection{Weakly Supervised Segmentation Comparison}
\label{sec:weak_comp}

Block annotation can be considered a form of weakly supervised annotation where
a small number of pixels in an image are labelled. Representative works in this
area include \cite{lin2016scribblesup,bearman2016what,pathak2014fully,
pap2015weakly,dai2015boxsup}. Table 3 of \cite{lin2016scribblesup} is replicated
here (table \ref{table:weak_table}) for reference, and extended with our
results. All existing results show performance with a VGG-16 based model. We
train a MobileNet based model which has been shown to achieve similar
performance to VGG-16 (71.8\% vs 71.5\% Top-1 accuracy on ImageNet) while
requiring fewer computational resources ~\cite{mobilenet, mobilenetv2}.
Our fully-supervised implementation pretrained on ImageNet achieves 69.6\% mIOU
on Pascal VOC 2012 \cite{everingham10pascal}; in comparison, the reference
DeepLab-VGG16 model achieves 68.7\% mIOU ~\cite{deeplabv2} and the
re-implementation in ~\cite{lin2016scribblesup} achieves 68.5\% mIOU.

\begin{table}[th!]
 \small
 \begin{tabularx}{\linewidth}{|X||X|X|}
   \hline
   \makecell[l]{Method} &
   \makecell[l]{Annotations} &
   \makecell[l]{mIOU (\%)} \\
   \hline
   \hline
   \makecell[l]{MIL-FCN \cite{pathak2014fully}} & Image-level & 25.1 \\
   \makecell[l]{WSSL \cite{pap2015weakly}} & Image-level & 38.2 \\
   \makecell[l]{point sup. \cite{bearman2016what}} & Point & 46.1 \\
   \makecell[l]{ScribbleSup \cite{lin2016scribblesup}} & Point & 51.6 \\
   \makecell[l]{WSSL \cite{pap2015weakly}} & Box & 60.6 \\
   \makecell[l]{BoxSup \cite{dai2015boxsup}} & Box & 62.0 \\
   \makecell[l]{ScribbleSup \cite{lin2016scribblesup}} & Scribble & 63.1 \\
   \hline
   \makecell[l]{\!\!\textbf{Ours:} Block-1\%\!\!\!}& Pixel-level Block & 61.2 \\
   \makecell[l]{\!\!\textbf{Ours:} Block-5\%\!\!\!}& Pixel-level Block & 67.6 \\
   \makecell[l]{\!\!\textbf{Ours:} Block-12\%\!\!\!\!\!}& Pixel-level Block & 68.4 \\
   \hline
   \hline
   \makecell[l]{Full Supervision} & Pixel-level Image & 69.6 \\
   \hline
\end{tabularx}
\vspace{-1em} 
\caption{\small \textbf{Weakly-supervised segmentation performance}.
Evaluated on Pascal VOC 2012 validation set. Original table from
\cite{lin2016scribblesup}. Blocks (N\%) indicates N\% of image pixels (N
pseudo-checkerboard blocks) are labelled. } \label{table:weak_table}
\vspace{-2mm}
\end{table}

\begin{table}[th!]
 \small
 \begin{tabularx}{\linewidth}{|c|l|l||X|}
   \hline
   \makecell[l]{Cityscapes} &
   \makecell[l]{\!\!\textbf{Ours:} Block\\\!\!(7 min)\!\!}&
   \makecell[l]{\!\!Coarse\!\! \\\!\!(7 min \cite{cityscapes})} &
   \makecell[l]{\!\!Full Supervision\!\!\\ \!\!(90 min \cite{cityscapes})} \\

   \hline
   \makecell[l]{mIOU (\%)} & \textbf{72.1} & 68.8 & 77.7  \\
   \hline
   \hline
   \makecell[l]{Pascal} &
   \makecell[l]{\!\!\textbf{Ours:} Block\\\!\!(25 sec)\!\!} &
   \makecell[l]{\!\!Scribbles\\\!\!(25 sec \cite{lin2016scribblesup})} &
   \makecell[l]{\!\!Full Supervision\!\!\\ \!\!(4 min \cite{pascalcontext})} \\

   \hline
   \makecell[l]{mIOU (\%)} & \textbf{67.2}  & 63.1 \cite{lin2016scribblesup} &
   69.6 \\

   \hline
\end{tabularx}
\vspace{-1em} \caption{\small \textbf{Weakly-supervised segmentation performance given
equal annotation time}.  
For time comparison of
scribbles against other methods, please refer to \cite{lin2016scribblesup}.}
\label{table:weakly_sup} \vspace{-2mm}
\end{table}

\mypara{Performance Comparison.} 
With only 1\% of the pixels annotated, block annotation achieves comparable
performance to existing weak supervision methods. Based on our results in
section \ref{sec:block_quality}, the cost of annotation for 1\% of pixels with
blocks will be 100$\times$ less than the cost of full-image annotation.
Increasing the budget to 5\%-12\% significantly increases performance. With 12\%
of pixels annotated with blocks, the segmentation performance (error) is within
98\% (4\%) of segmentation performance (error) with 100\% of pixels annotated.

Note that block annotations can be directly transformed into gold-standard fully
dense annotations by simply gathering more block annotations within an image.
This is not feasible with other annotations such as point clicks, scribbles, and
bounding boxes. Furthermore, in section \ref{sec:block_to_dense}, we demonstrate
a method to transform block annotations into dense
annotations without any additional human effort.

\mypara{Equal Annotation Time Comparison.} Given equal annotation time, block
annotation significantly outperforms coarse and scribble annotations by
$\sim$3-4\% mIOU (table \ref{table:weakly_sup}). On Pascal, 97\% of
full-supervision mIOU is achieved with 1/10 annotation time.  We convert
annotation time to number of annotated blocks as follows.  Block annotation may
use up to 2.2$\times$ the time of full-image annotation.  Given an image divided
into 100 blocks, an annotation time of $T$ leads to $\frac{T}{0.022F}$ (eq. 5)
blocks annotated, where $F$ is the full-image annotation time.

\section{Block-Inpainting Annotations \label{sec:block_to_dense}}

Although block annotations are useful for learning semantic segmentation, the
full structure of images is required for many applications. Understanding the
spatial context or affordance relationships \cite{cocostuff, hassanin2018visual}
between classes relies on understanding the role of each pixel in an image.
Shape-based retrieval, object counting \cite{lempitsky2010learning}, or
co-occurrence relationships \cite{mivcuvslik2009semantic} also depend on a
global understanding of the image. The naive approach to recover pixel-level
labels is to use automatic segmentation to predict labels. However, this does
not leverage existing annotations to improve the quality of predicted labels. In
section \ref{sec:block_to_dense_model}, we propose a method to inpaint
block-annotated images by using annotated blocks as context.  In section
\ref{sec:block_to_dense_eval}, we examine the quality of these inpainted
annotations. 

\subsection{Block-Inpainting Model}
\label{sec:block_to_dense_model}

The goal of the block-inpainting model is to inpaint labels for unannotated
blocks given the labels for annotated blocks in an image. For full
implementation details and ablation studies, please refer to the supplemental.

\mypara{Architecture.} The block-inpainting model is based on DeepLabv3+. The
input layer is modified so that the RGB image, $I \in \mathbb{R}^{h \times w
\times 3}$, is concatenated with multichannel ``hint''
(ala~\cite{zhang2017real}) of 1-0 class labels $W \in \mathbb{R}^{h \times w
\times K}$ where $K$ is the number of classes. At inference time, the hint
contains known labels for the annotated blocks of an image which serve as
context for the inpainting task. Hidden layers are augmented with dropout
which will be used to control quality by estimating epistemic
uncertainty~\cite{gal2015bayesian, gal2016dropout}.

\mypara{Estimating Uncertainty.} Inpainting fills all missing regions without
considering the trade off between quantity and quality. Existing datasets have
high-quality annotations for 92-94\% of pixels ~\cite{ade20k, cocostuff}.
Therefore, we modify our network to produce uncertainty estimates which allow us
to explicitly control this trade off. The uncertainty of predictions is
correlated with incorrect predictions~\cite{leibig2017leveraging,
laptev2017time}.  Uncertainty is computed by activating dropout at inference
time.  The predictions are averaged over the $g$ trials giving us $U \in
\mathbb{R}^{h \times w}$, a matrix of uncertainty estimates per image. We take
the sample standard deviation corresponding to the predicted class for each pixel to be
the uncertainty. For each pixel $(i,j)$, the mean softmax vector over $g$
trials is:

\begin{equation}
  \boldsymbol{\mu}^{(i,j)} = \frac{\sum\limits_{t=1}^g \textbf{p}^{(i,j)}(y|I,
W)}{g}
\end{equation}
where $\textbf{p}(y|I, W) \in \mathbb{R}^{K}$ is the softmax output of the
network. The corresponding uncertainty vector is:
\begin{equation}
  \textbf{U}'^{(i,j)} = \sqrt{\frac{\sum\limits_{t=1}^g (\textbf{p}^{(i,j)}(y|I,
  W) -
  \boldsymbol{\mu}^{(i,j)})^2}{g - 1}}\\
\end{equation}
Thus, the uncertainty for each pixel $(i,j)$ is:
\begin{equation}
  U^{(i,j)} = {U}'^{(i,j)}_{m} \text{, where }m=\argmax\limits_k
  {\mu}^{(i,j)}_{k}\\
\end{equation}

\mypara{Training.} Block annotations serve both as hints and targets. This means
that no additional data (or human annotation effort) is required to train the
block-inpainting model. For our experiments, we use (synthetically generated)
Block-50\% annotations.  For each image, half of annotated blocks are randomly
selected online at training time to be hints. All of the annotated blocks are
used as targets. This encourages the network to ``copy-paste'' hints in the final
output while leveraging the hints as context to inpaint labels for regions where
hints are not provided.

\subsection{Evaluation}
\label{sec:block_to_dense_eval}

\begin{figure}[t!]
  \centering
  \begin{subfigure}[t]{0.47\linewidth}
      \includegraphics[width=\linewidth]{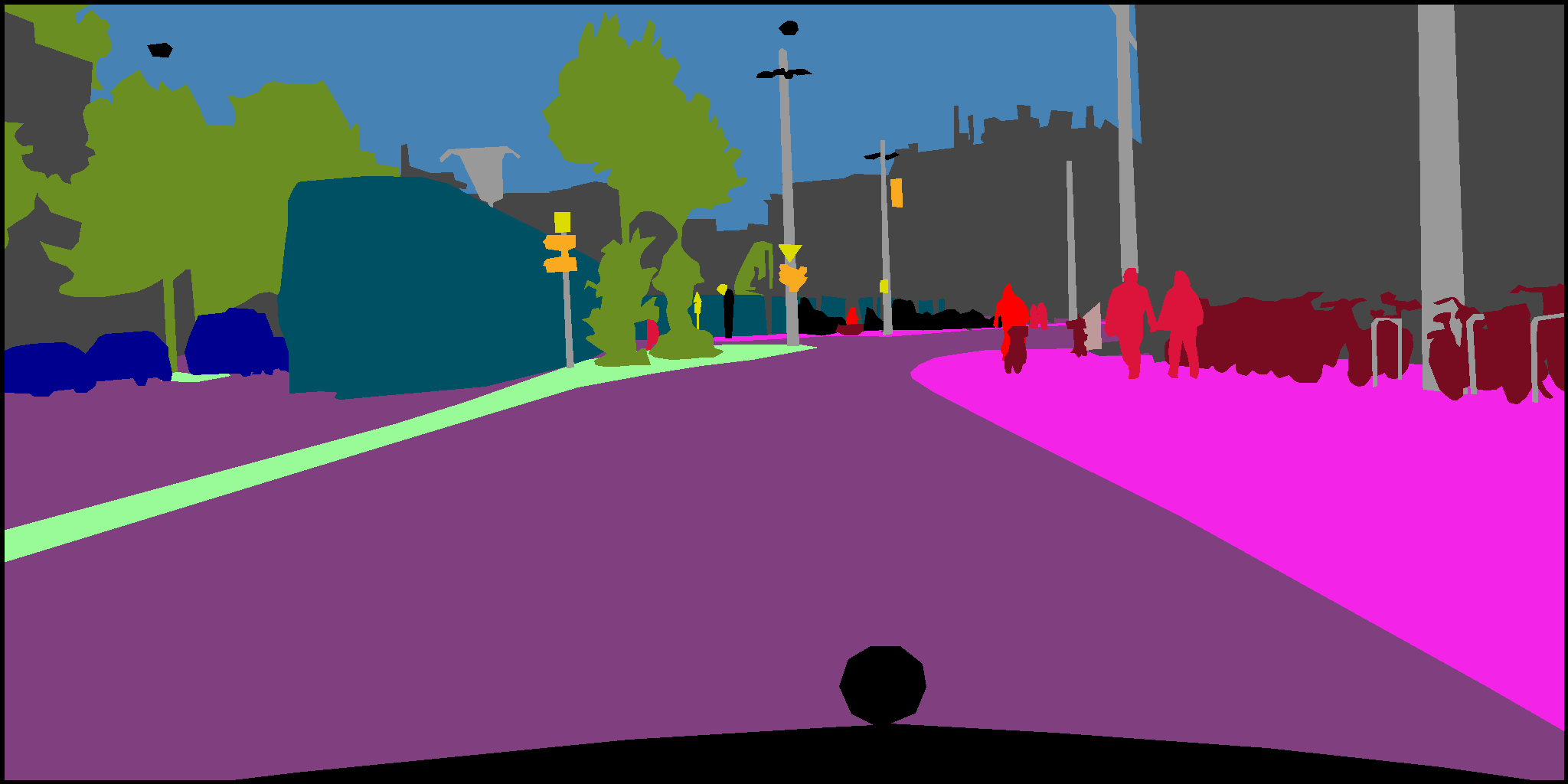}
      \caption{\small Full human labels}
  \end{subfigure}
  \hfill
  \begin{subfigure}[t]{0.47\linewidth}
      \includegraphics[width=\linewidth]{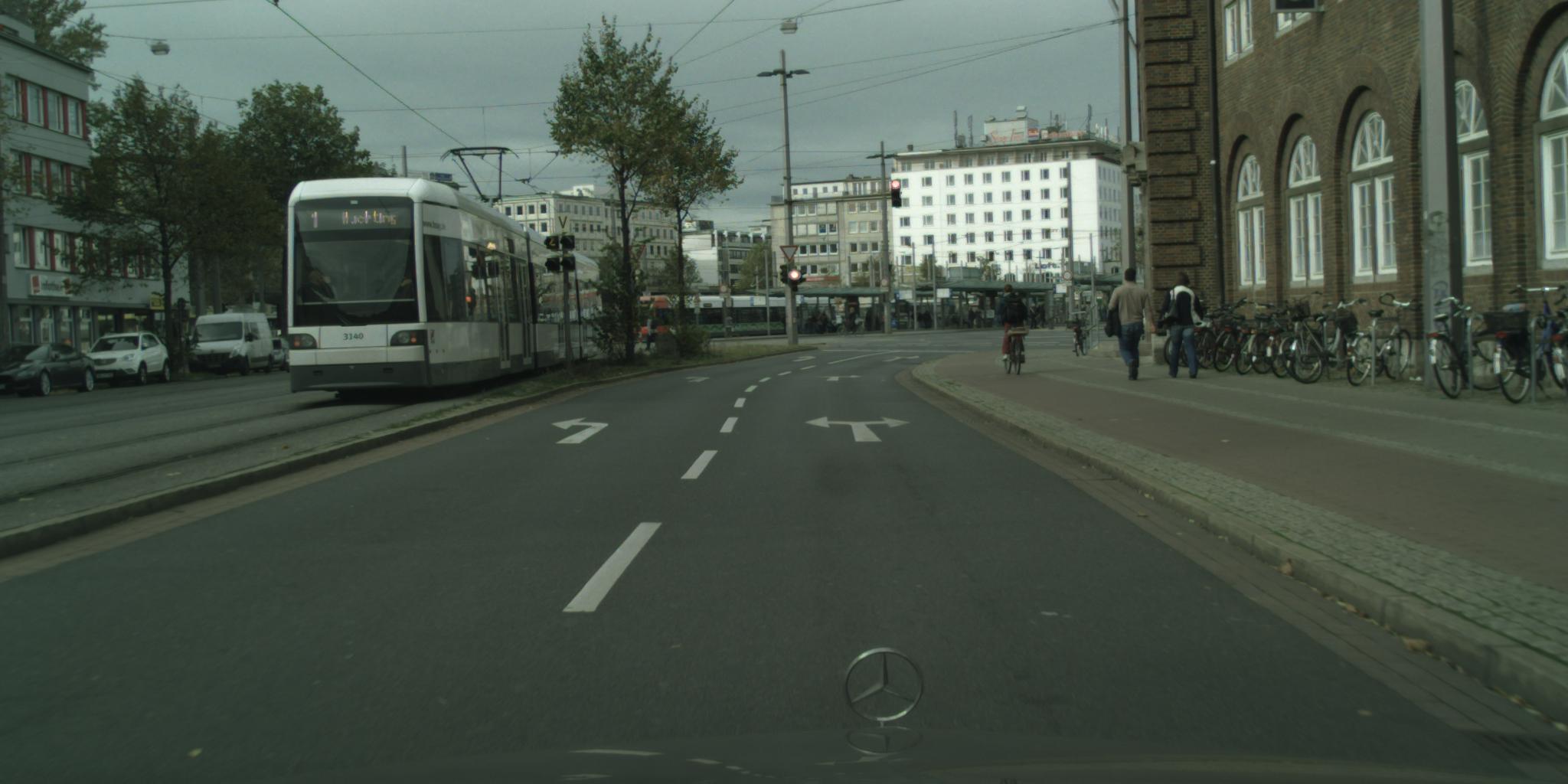}
      \caption{\small Original image}
  \end{subfigure}
  \begin{subfigure}[t]{0.47\linewidth}
      \includegraphics[width=\linewidth]{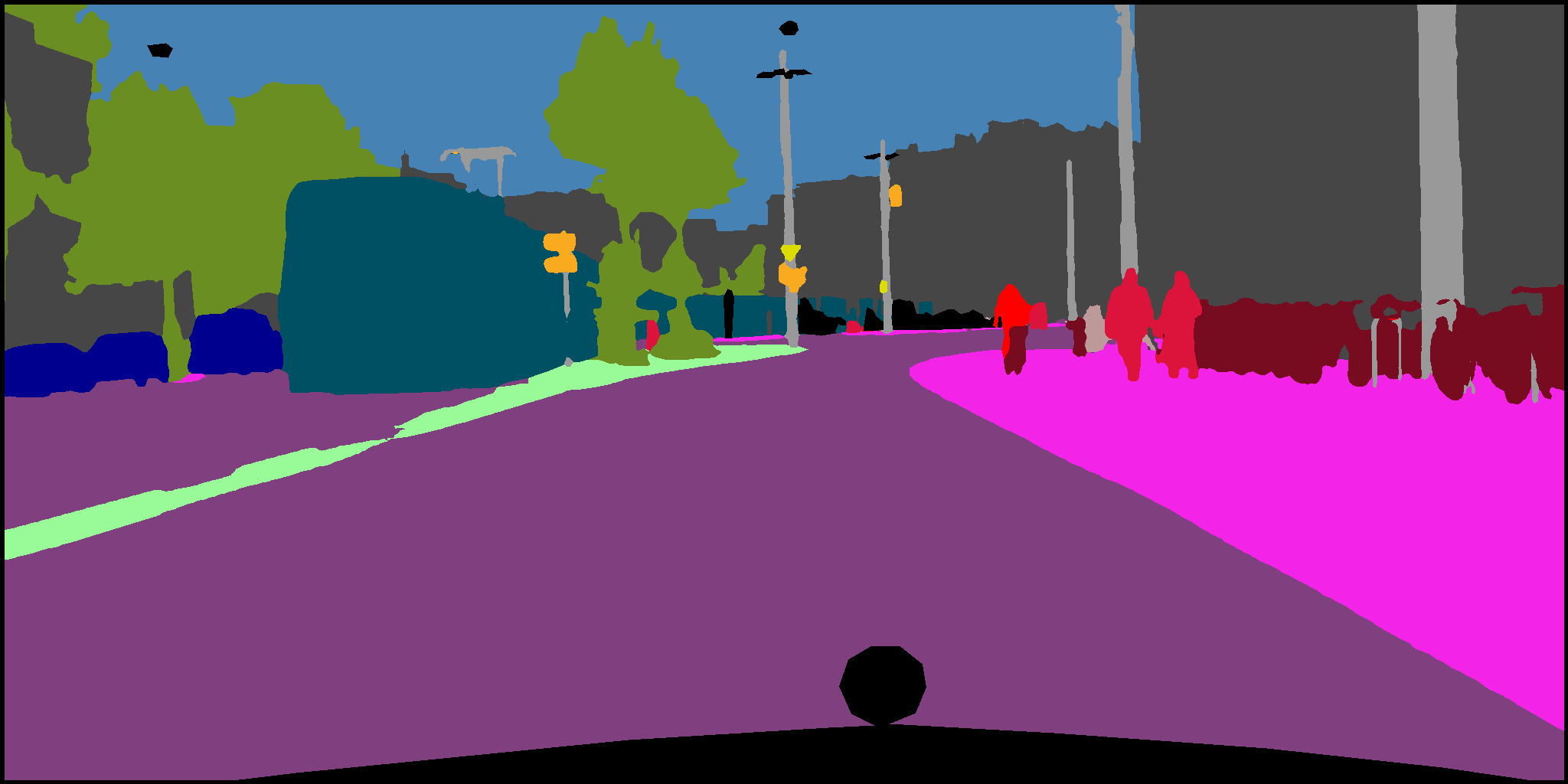}
      \caption{\small Inpainted labels (all)}
  \end{subfigure}
  \hfill
  \begin{subfigure}[t]{0.47\linewidth}
      \includegraphics[width=\linewidth]{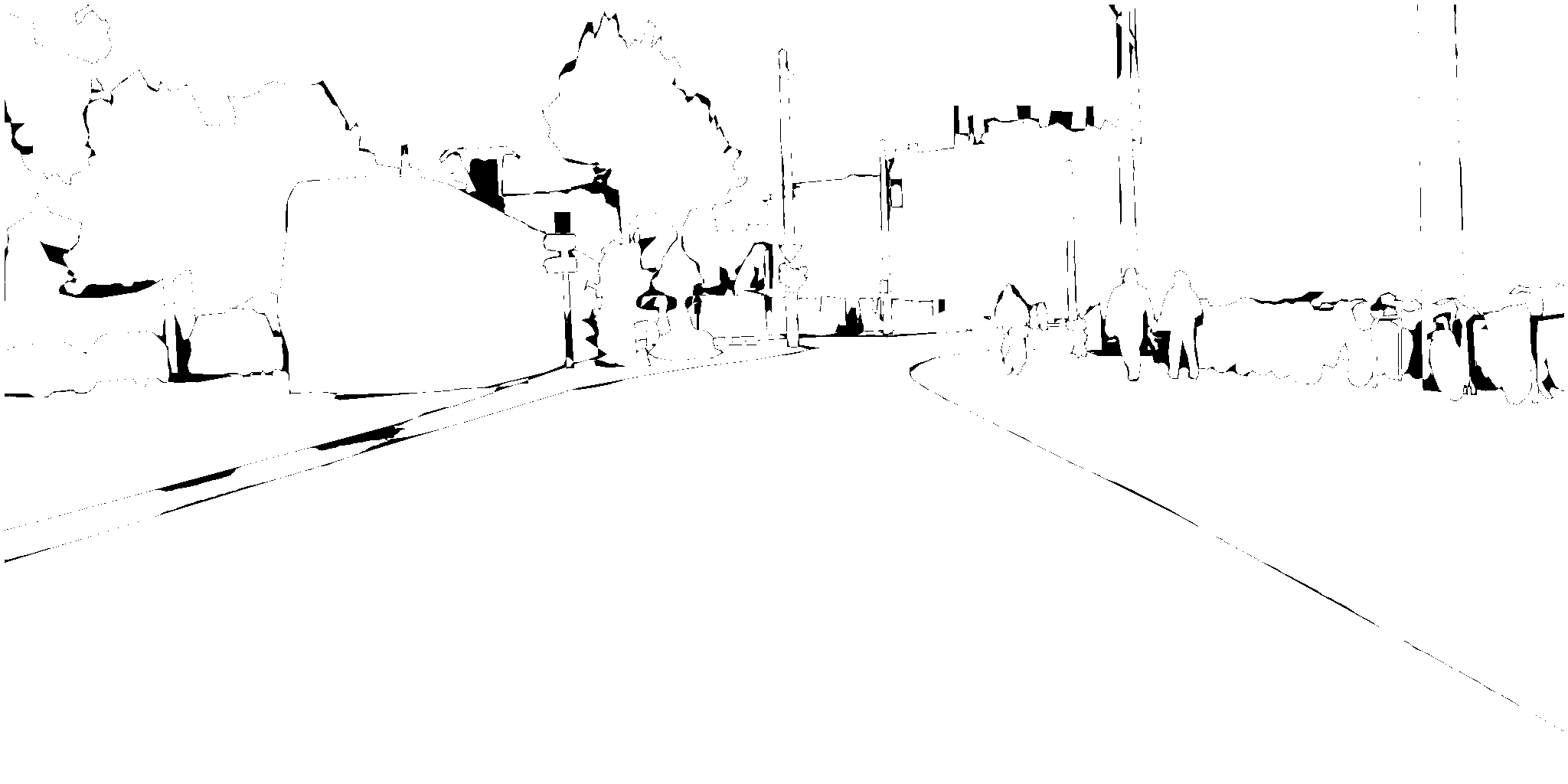}
      \caption{\small Label agreement (white)}
  \end{subfigure}
  \begin{subfigure}[t]{0.47\linewidth}
      \includegraphics[width=\linewidth]{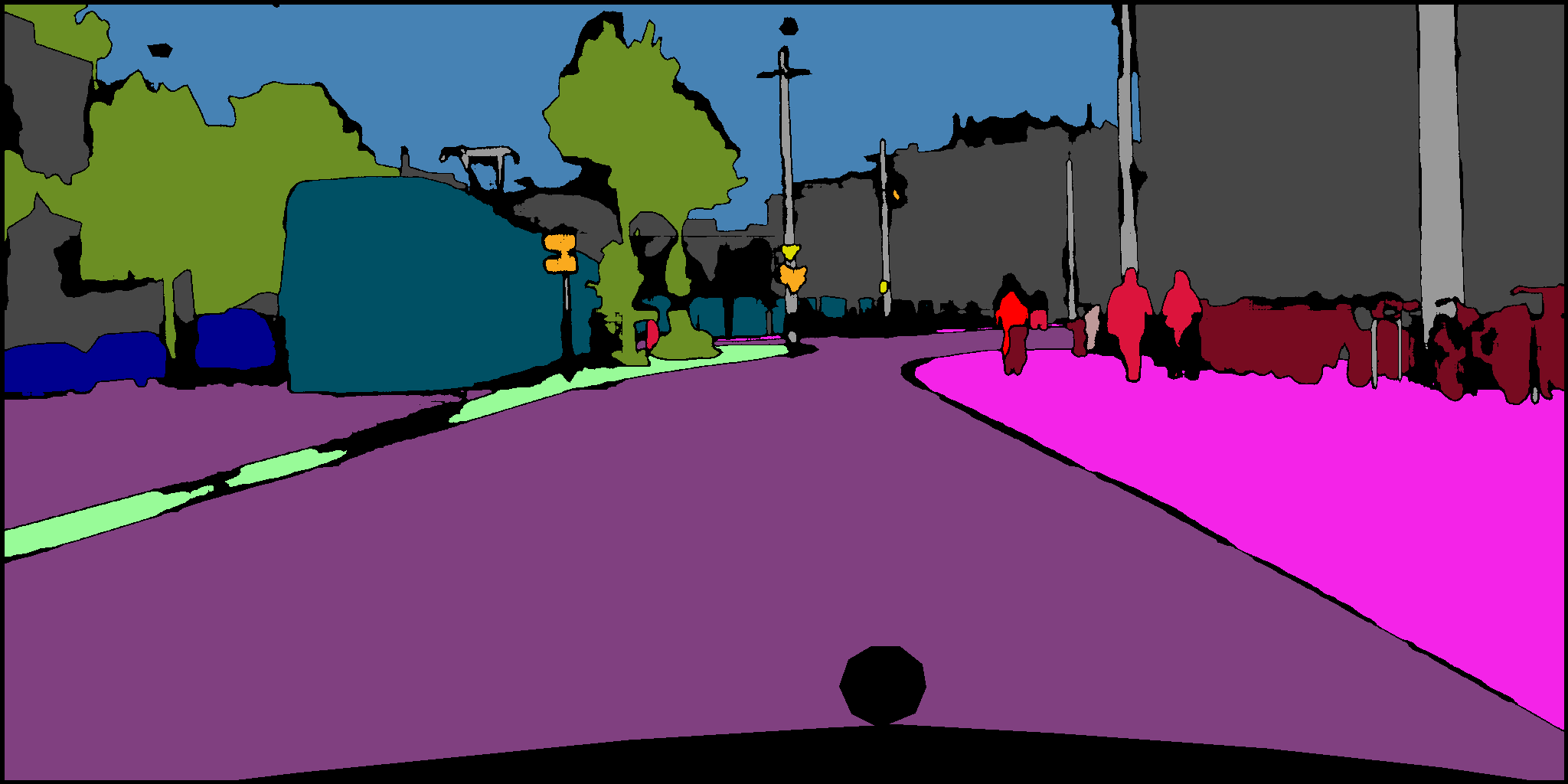}
      \caption{\small Inpainted labels ($<$20\% relative uncertainty)}
  \end{subfigure}
  \hfill
  \begin{subfigure}[t]{0.47\linewidth}
      \includegraphics[width=\linewidth]{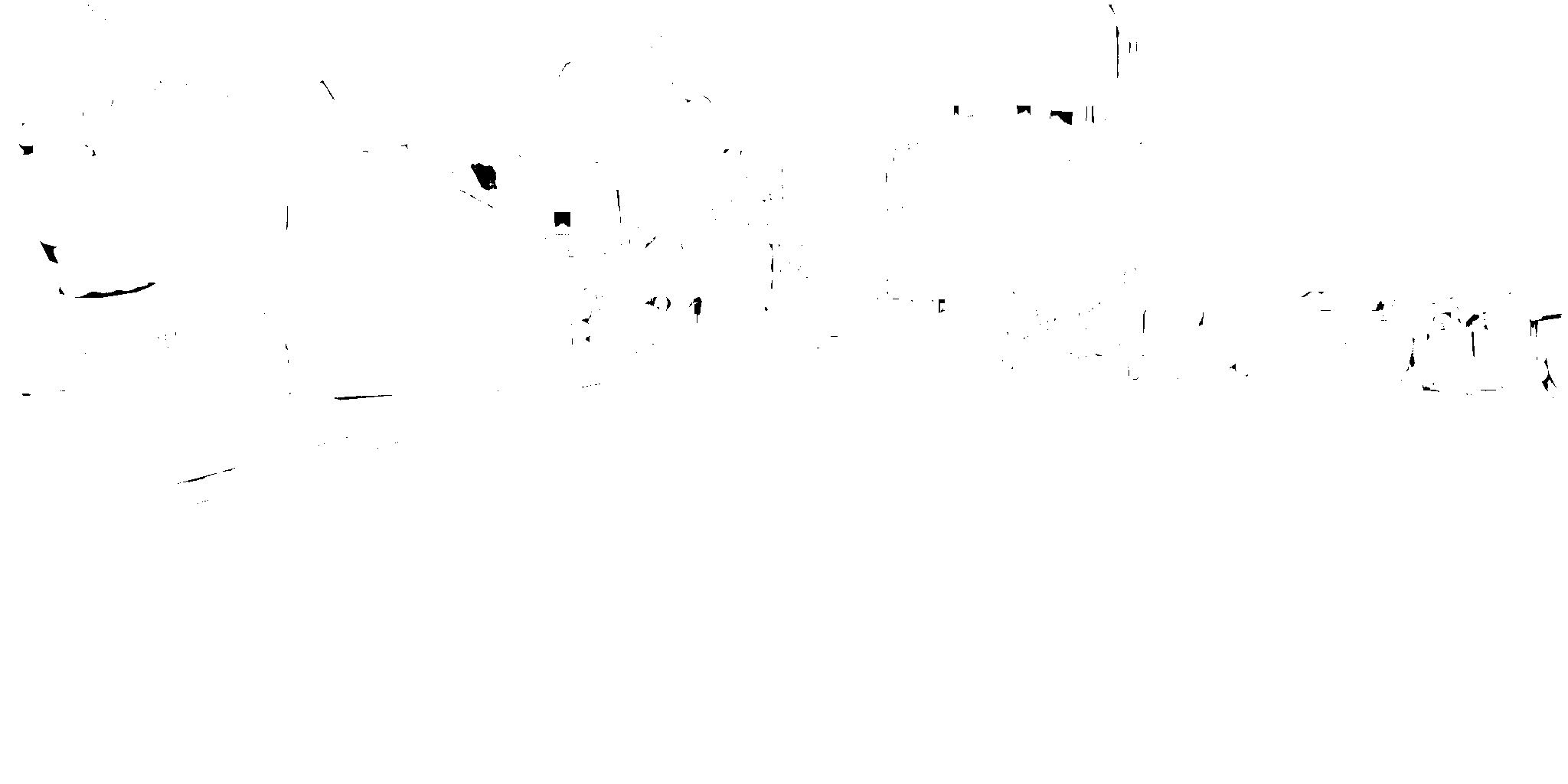}
      \caption{\small Label agreement (white)}
  \end{subfigure}
\vspace{-1em}
\caption{\small \textbf{Block-inpainted labels.} Example of human labels vs human
    Block-50\% + inpainted labels. Void labels are masked out. }
    \label{fig:GT_vs_low_uncert}
\vspace{-2mm}
\end{figure}

\paragraph{Quality of Inpainted Labels.} How good are inpainted labels?  We
compare labels produced by the block-inpainting network with low $U^{(i,j)}$
against the known human labels in Cityscapes and ADE20K.  \emph{The
  block-inpainting model produces labels whose human-agreement is competitive
with that achieved by human annotators.} We inpaint Block-50\% annotations in
this experiment. At a relative uncertainty threshold of
0.2 (0.4) on Cityscapes (ADE20K), over 94\% of the pixels are labelled. The mean
pixel agreement is 99.8\% (98.7\%) and the class-balanced error rate is 3.1\%
(28\%). Previous work show that human label agreement across annotators is
66.8\% to 73.6\% while annotator self-agreement is 82.4\% to
97.0\%~\cite{ade20k, cocostuff}.  Human annotators fail to agree in non-trivial
fashion -- ~\cite{ade20k} shows that annotator self-agreement fails in three
ways: variations in complex boundaries (32\%), incorrect naming of ambiguous
classes (34\%), and failure to segment small objects (34\%).  In figure
\ref{fig:GT_vs_low_uncert}, a visualization of labels generated by the
block-inpainting model is shown. The number of pixel disagreements decreases
with a higher uncertainty threshold.

\mypara{Block Inpainting vs Automatic Segmentation.} Consider a scenario in
which a small number of pixels in a dataset are annotated, and the remainder are
automatically labelled to produce dense annotations. Why should block inpainting
be used instead of automatic segmentation? \emph{Full pixel-level labels
produced by block inpainting are superior to automatic segmentation.} On
Cityscapes, automatic segmentation achieves 78\% validation mIOU while block inpainting
Block-50\% annotations achieves 92\% validation mIOU. With Block-12\% annotations,
automatic segmentation achieves 75\% validation mIOU while block inpainting achieves 82\%
validation mIOU.

\mypara{Block Selection vs Block-Inpainting Quality}
How does the checkerboard pattern compare to other block selection strategies as
hints to the block-inpainting model? Intuitively, it is easier to infer labels
for pixels that are close to pixels with known labels than for pixels that are
further away. Consider a scenario in which every other pixel in an image is
annotated. Reasonably good labels for the unannotated pixels can be inferred
with a simple nearest-neighbors algorithm. In practice, it is impossible to
precisely annotate single pixels in an image. However, we can approximate the
same properties of labelling every other pixel by labelling every other block
instead (i.e., a checkerboard pattern).

In table \ref{table:anno_hints_miou}, we show the block-inpainting model mIOU
when different types of hints are given. The rightmost column (``every other
pixel'') is not feasible to collect in practice. Checkerboard annotations
outperform random block annotations even though the network is trained to expect
random block hints. Providing only boundary annotations within each block (i.e.
annotating pixels within 10 pixels of each boundary in each block) allows the
network to achieve nearly the same performance as full block hints. This
suggests that the most informative pixels for the block-inpainting model are
those near a boundary.

\begin{table}[t!]
 \small
 \begin{tabularx}{\linewidth}{|c|l|l|l|l|X|}
   \hline
   \makecell[l]{} & 
   \makecell[l]{\!\!None\!\!\!\!\!} &
   \makecell[l]{\!\!Random\\\!\!(Bndy)} &
   \makecell[l]{\!\!Random\\\!\!(Full)} &
   \makecell[l]{\!\!Checker\\\!\!(10x10)\!\!} &
   \makecell[l]{\!\!Every oth.\\\!\!pixel\!\!} \\
   \hline
   \makecell[l]{\!\!Rel. mIOU\!\!} & 0.77 & 0.90 & 0.92 &  0.95 & 1.0 \\
   \hline
\end{tabularx}
\vspace{-1em} \caption{\small \textbf{Block-inpainting with different
types of hints.}  ``Every other pixel'' annotations are infeasible in practice.
Relative performance of hints with respect to ``every other pixel'' hints is
shown. Checkerboard blocks outperform no hints, random blocks (only boundaries
within blocks), and random blocks (full blocks).  }
\label{table:anno_hints_miou} \vspace{-2mm}
\end{table}

\section{Conclusion}

In this paper we have introduced block annotation as a replacement for
traditional full-image annotation with public crowdworkers. For semantic
segmentation, Block-12\% offers strong performance at 1/8th of the monetary
cost. Block-5\% offers competitive weakly-supervised performance at equal
annotation time to existing methods. For optimal semantic segmentation
performance, or to recover global structure with inpainting, Block-50\% should
be utilized.

There are many directions for future work. Our crowdworker tasks are similar to
full-image annotation tasks so it may be possible to improve the gains with more
exploration and development of boundary marking algorithms. We have explored
some block patterns and further exploration may reveal even better trade-offs
between annotation quality, cost and image variety. Another interesting
direction is acquiring instance-level anntoations by merging segments across
block boundaries. Active learning can be used to select blocks of rare classes,
and workers can be assigned blocks so that annotation difficulty matches worker
skill.

\paragraph{Acknowledgements.}
We acknowledge support from Google, NSF (CHS-1617861 and CHS-1513967), PERISCOPE
MURI Contract \#N00014-17-1-2699, and NSERC (PGS-D). We thank the reviewers for
their constructive comments. We appreciate the efforts of MTurk workers who
participated in our user studies.

\newpage
{\small
\bibliographystyle{ieee_fullname}
\bibliography{paper}
}

\end{document}